\documentclass[letterpaper, 10pt, conference]{ieeeconf}  

\IEEEoverridecommandlockouts                              

\usepackage{url}

\usepackage[utf8]{inputenc}
\usepackage{graphicx}
\graphicspath{{figures/}}
\usepackage{epsfig}

\usepackage{amsmath}
\usepackage{amsthm}
\usepackage{amssymb}
\usepackage{amsfonts}
\usepackage{comment}
\usepackage{bbm}
\usepackage{subcaption}
\usepackage{tikz}
\usepackage{tkz-euclide}

\usetkzobj{all}
\usetikzlibrary{arrows, shapes, decorations.markings, calc, patterns, angles, quotes, intersections, plotmarks}
\usepackage{aircraftshapes}

\usepackage{float}
\usepackage[ruled,vlined]{algorithm2e}
\newcommand{\spr}[1]{{\color{red} #1}}

\newtheorem{definition}{Definition}[section]
\newtheorem{theorem}{Theorem}[section]
\usepackage{mathtools}

\usepackage{hyperref}
\usepackage{cleveref}

\title{\LARGE \bf
Multi-Robot Target Search using Probabilistic Consensus on \\ Discrete Markov Chains
}
\author{Aniket Shirsat, Karthik Elamvazhuthi, and Spring Berman
\thanks{This work was supported by ONR Young Investigator Award N00014-16-1-2605 and by the Arizona State University Global Security Initiative.}
\thanks{Aniket Shirsat and Spring Berman are with the School for Engineering of Matter, Transport and Energy, Arizona State University, Tempe, AZ, 85287 USA {\tt\small \{ashirsat, Spring.Berman\}@asu.edu}.}
\thanks{Karthik Elamvazhuthi is with the Department of Mathematics, University of California, Los Angeles, CA, 90095 USA {\tt\small karthikevaz@math.ucla.edu}.}
}

\begin{document}
\maketitle
\pagestyle{empty}

\begin{abstract}
In this paper, we propose a probabilistic consensus-based multi-robot search strategy that is robust to communication link failures, and thus is suitable for disaster affected areas. The robots, capable of only local communication, explore a bounded environment according to a random walk modeled by a discrete-time discrete-state (DTDS) Markov chain and exchange information with neighboring robots, resulting in a time-varying communication network topology. The proposed strategy is proved to achieve consensus, here defined as agreement on the presence of a static target, with no assumptions on the connectivity of the communication network. Using numerical simulations, we investigate the effect of the robot population size, domain size, and information uncertainty on the consensus time statistics under this scheme. We also validate our theoretical results with 3D physics-based simulations in Gazebo. The simulations demonstrate that all robots achieve consensus in finite time with the proposed search strategy over a range of robot densities in the environment.
\end{abstract}
\section{INTRODUCTION}
Disaster areas, such as regions affected by earthquakes and floods, experience great disruption to communication and power infrastructure. This presents challenges in coordinating searches for survivors and dispersing relief teams to those locations.
Teams of mobile robots have proved to be useful for exploring and mapping environments in disaster response scenarios \cite{michael2014collaborative,burgard2005coordinated,nagatani2013emergency}. However, such robots are subject to constraints on the payloads that they can carry, including power sources, sensors, embedded processors, actuators, and communication devices for transmitting information to other agents and/or to a  command center. In addition, many  multi-robot control strategies rely on a communication network for coordination.
Centralized exploration strategies like \cite{simmons2000coordination} rely on constant communication between agents and a central node. However, these strategies do not scale well with the number of agents, since the communication bandwidth becomes a bottleneck with increasing agent population size. Moreover, such strategies  suffer from a single point of failure, i.e., a disruption to the central node causes loss of communication for all the agents. 

\begin{figure}[t]
    \centering
    \includegraphics[width=0.35\textwidth]{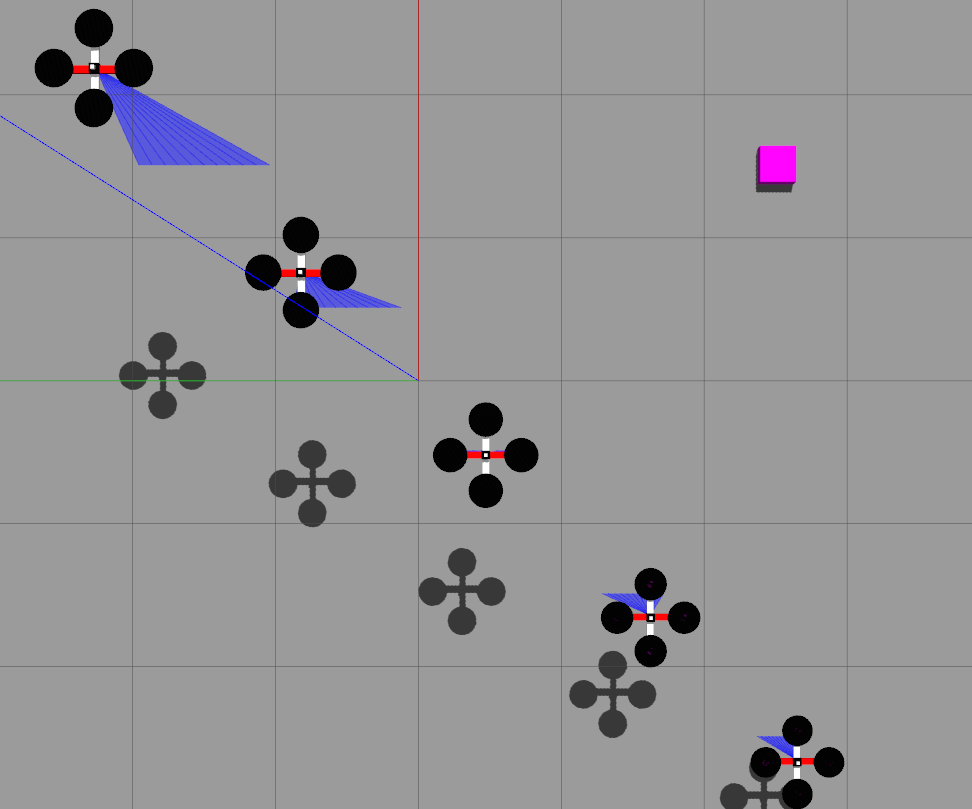}
    \caption{Overhead view of problem scenario, simulated in Gazebo 9 \cite{koenig2004design}. Multiple aerial robots, flying at different heights, search for a target represented by the magenta box using a Markov chain motion model.}
    \label{fig:gazebo_environment}
\end{figure}

These drawbacks can be overcome by employing decentralized exploration strategies that involve only local communication between agents. However, communication can become unreliable as the number of agents increases \cite{howard2006experiments}, and the connectivity of the communication network may be disrupted in some applications by the environment \cite{husain2013mapping} or by the movement of agents outside of communication range. 
Decentralized multi-agent control strategies that employ communication networks often require the agents to reach {\it consensus} on a particular variable.
Achieving consensus is the problem of arriving at a common output variable or global property from measurements by distributed agents with local communication, without the need for a supervisory agent (leader or central processor) \cite{spanos2005dynamic}. 
Consensus problems have been studied in the cases of static or fixed network topologies \cite{ren2007information} and dynamic or switching network topologies \cite{spanos2005dynamic}, directed and undirected communication graphs \cite{ren2004consensus}, random networks \cite{mesbahi2010graph}, and mobile networks with communication delays \cite{olfati2004consensus}. 
Consensus algorithms for multi-robot rendezvous, e.g. \cite{parasuraman2018multipoint,yu2020synthesis,fox2006distributed,vincent2008distributed}, are an example of such a strategy on a dynamic network. The robot controllers drive the robots to meet at a common location in order to enable their information exchange via local communication. However, such strategies restrict exploration since the robots must aggregate at a common location. Distributed consensus for merging individual agents' information
has been previously used for multi-agent search, e.g. \cite{hu2012multiagent}; however, it requires a connected communication network. 
Although random mobility models are commonly used in multi-robot exploration, e.g. \cite{martinez2012brownian,wagner1998robotic,winfield2000distributed}, few works consider consensus problems for agents that perform probabilistic search strategies, and thus have randomly time-varying communication networks.

To address this problem, we present and analyze a probabilistic multi-agent search strategy that is based on a distributed consensus protocol. The proposed strategy is decentralized and asynchronous and relies on only limited communication among agents. Thus, it can be employed in applications, such as disaster response scenarios, where it is infeasible to maintain a connected communication network, rendezvous, or communicate with a central node.
The agents 
move according to a discrete-time discrete-state (DTDS) Markov chain model on a finite spatial grid, as illustrated in \autoref{fig:gazebo_environment}.
We consider only static features here, which represent persistent characteristics of the target(s) that the agents are searching for in the environment. 
The main contributions of this paper are the following:
\begin{enumerate}
\item 
We prove that agents with a DTDS Markov motion model and local communication will achieve consensus, in an \textit{almost sure} sense, on the presence of a static feature of interest in a bounded environment.
\item Our proof does not require the assumption that the agent communication network remain connected over a non-zero finite time interval, as assumed in \cite{RageshTRO2020} for a similar consensus problem over a time-varying network.\footnote{This assumption implies the existence of a uniform upper bound on the interval between successive meeting times of any two agents, which is not guaranteed for agents that evolve stochastically on a finite connected state space.}
\end{enumerate} 
We validate our theoretical results with  Monte Carlo simulations in MATLAB and with 3D physics simulations performed in Gazebo 9 \cite{koenig2004design} using the Robot Operating System (ROS). From the simulation results, we empirically characterize the dependence of the expected time until consensus on the number of agents, the grid size, and the agent density, which can be used to guide the selection of the number of agents to search a given environment.

The remainder of the paper is organized as follows. Section \ref{sec:ProbStatement} presents the problem formulation, and Section \ref{sec:DSMC} describes the probabilistic motion model of the agents. Section \ref{sec:ConsensusBasedSearch} proves that all agents will reach consensus on the presence of the feature under our stochastic search strategy. Section \ref{sec:Sim} presents example implementations of our strategy in numerical and 3D physics simulations and discusses the results. Section \ref{sec:ConcFutre} concludes and suggests future work.

\section{PROBLEM STATEMENT} \label{sec:ProbStatement}
We consider an unknown, bounded environment that contains a finite, non-zero number of static features of interest, indexed by the set $\mathcal{I} \subset \mathbb{Z}_+$, where $\mathbb{Z}_+$ is the set of positive integers.
A set of $N$ agents, indexed by the set $\mathcal{N} = \{1,2,...,N\}$, explore the environment using a random walk strategy. 
We assume that each agent can localize itself in the environment and can detect a feature within its sensing range. When an agent $a \in \mathcal{N}$ detects a feature at discrete time $k$, it associates a scalar {\it information state} $\xi_{a}[k] \in \mathbb{R}_{\geq 0}$ with its current position. 
The vector of information states for all agents at time $k$ is denoted by $\boldsymbol{\xi}[k]$. Defining $\mathcal{U}(0,1)$ as the uniform probability distribution on the interval $[0,1]$, the initial information state of each agent $a$ is specified {\it a priori} as $\xi_a[0] \sim \mathcal{U}(0,1)$. 
The agent can communicate its information state $\xi_{a}[k]$ at time $k$ to all agents within a disc of radius $r_{comm} \in (0,\delta]$, where $\delta$ is the maximum communication radius. We define these agents as the set of {\it neighbors} of agent $a$ at time $k$, denoted by $\mathcal{N}_k^a$.
In addition, we assume that the agents can avoid  obstacles during their exploration. 
Since the agents are constantly moving, the set of agents with which they can communicate changes over time. The time evolution of this communication network is determined by the random walks of the agents throughout the bounded environment. This approach uses low communication bandwidth, since each agent only transmits a scalar value associated with each feature that it detects.

\begin{figure}[t]
    \centering
        \begin{tikzpicture}[scale=0.7]
            \draw [step=1cm, thin, gray!50] (0,0) grid (10,10);
            \filldraw[color=black]  (5,5) circle (0.1);
            \node[right] (i) at (4.8,4.5) {$m=Z^1_k=Z^2_k$};
	        \node [quadcopter top,fill=white,draw=red,minimum width=0.5cm, rotate=45] at (3.0,1.0) {};
	        \node[fill=red,regular polygon, regular polygon sides=3,inner sep=2.5pt] at (2.0,1) {};
	        \node[above] at (1.5,0.0) {$Z^1_0=i$}; 
	        \node[star,star points=8, fill=magenta,draw=black] at (8.1,7.9){};
            \node [quadcopter top,fill=white,draw=orange,minimum width=0.5cm, rotate=45] at (5,9) {};
            \node[fill=orange,regular polygon, regular polygon sides=3,inner sep=2.5pt] at (5,8) {}; 
        	\node[above] at (4.2,7) {$Z^2_0 = j$};
        	\draw[->,color=red, thick] (2,1) -- (1,1);
        	\draw[->,color=red, thick] (1,1) -- (1,2);
        	\draw[->,color=red, thick] (1,2) -- (2,2); 
        	\draw[->,color=red, thick] (2,2) -- (2,3);
        	\draw[->,color=red, thick] (2,3) -- (2,4);
        	\draw[->,color=red, thick] (2,4) -- (3,4);
        	\draw[->,color=red, thick] (3,4) -- (4,4);
        	\draw[->,color=red, thick] (4,4) -- (4,5);
        	\draw[->,color=red, thick] (4,5) -- (5,5);
        	\draw[->,color=orange, thick] (5,8) -- (6,8);
        	\draw[->,color=orange, thick] (6,8) -- (7,8);
        	\draw[->,color=orange, thick] (7,8) -- (8,8);
        	\draw[->,color=orange, thick] (8,8) -- (8,7);
        	\draw[->,color=orange, thick] (8,7) -- (7,7);
        	\draw[->,color=orange, thick] (7,7) -- (7,6);
        	\draw[->,color=orange, thick] (7,6) -- (6,6);
        	\draw[->,color=orange, thick] (6,6) -- (6,5);
        	\draw[->,color=orange, thick] (6,5) -- (5,5); 
        	\node [quadcopter top,fill=white,draw=black, minimum width=1.0cm, rotate=45] at (8,2) {};
            \draw [->, color=black, thick] (8,3) -- node[right] {up} (8,4);
            \draw [->, color=black, thick] (7,2) -- node[below]{left} (6,2);
            \draw [->, color=black, thick] (8,1) -- node[right]{down} (8,0);
            \draw [->, color=black, thick] (9,2) -- node[below]{right} (10,2);
        \end{tikzpicture}
        \caption{Illustration of our multi-agent search strategy, showing  sample paths for two quadrotors (orange and red) on a square grid. The quadrotors search the environment for a static target (the magenta star) as they perform a random walk on the grid.} 
        \label{fig:MotionStrategy_Visual}
\end{figure}
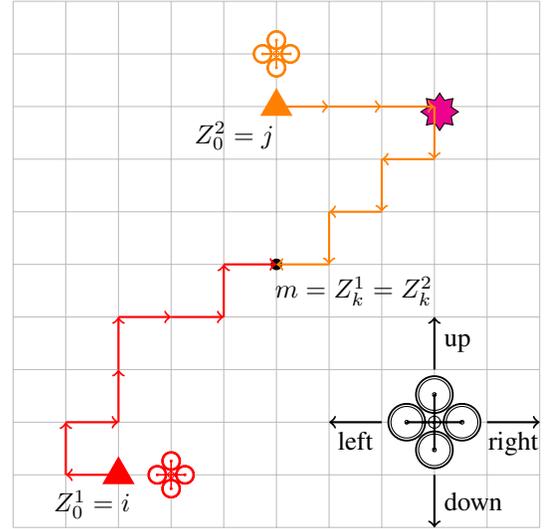

We discretize the environment, as shown in \autoref{fig:MotionStrategy_Visual}, into a square grid of nodes spaced at a distance $d$ apart. The set of nodes is denoted by $\mathcal{S} \subset \mathbb{Z}_+$. We define $S = | \mathcal{S} |$. Let $\mathcal{G}_{s} = (\mathcal{V}_{s}, \mathcal{E}_{s})$ be an undirected graph associated with this finite spatial grid, where $\mathcal{V}_{s}$ is the set of nodes and $\mathcal{E}_{s}$ is the set of edges $(i,j)$ that signify pairs of nodes $i,j \in \mathcal{V}_s$ between which agents can travel. We refer to these pairs of nodes as {\it neighboring nodes}.
Each agent performs a random walk on this grid, moving from its current node $i$ to a neighboring node $j$ at the next time step with transition probability $p_{ij}$. Let $Z_k^a \in \mathcal{S}$ be a random variable that represents the index of the node that an agent $a \in \mathcal{N}$ occupies at the discrete time $k$. For each agent $a$, the probability mass function $\pi_k \in \mathbb{R}^{1 \times S }$ of $Z_k^a$ evolves according to a DTDS Markov chain: 
\begin{equation}
    \pi_{k+1} = \pi_{k} \mathbf{P},
    \label{eqn:DiscreteStateTransitionDynamics}
\end{equation}
where the {\it state transition matrix} $\mathbf{P} \in \mathbb{R}^{S \times S}$ has elements $p_{i j} \in [0, 1]$ at row $i \in \mathcal{S}$ and column $j \in \mathcal{S}$.  

We assume that no prior information about possible search locations is available. To cover the search area uniformly, each agent is deployed from a random node on the spatial grid. These initial agent positions are chosen independently of one another and are identically distributed according to the probability mass function $\pi_{0}$, defined as a discrete uniform distribution over the set of nodes.
We define $\xi^{r} \in \mathbb{R}_{\geq 0}$ as a scalar {\it reference information state} that is associated with the set of nodes $\mathcal{Z}^{r} \subset \mathcal{S}$ from which an agent can detect a feature.
In this work, we consider environments with a single feature of interest.

We now define another graph that models the time-varying communication topology of the agents as they move along the spatial grid.
Let  $\mathcal{G}_c[k] = (\mathcal{V}_{c},\mathcal{E}_{c}[k])$ be an  undirected graph in which  $\mathcal{V}_{c} = \mathcal{N}$, the set of agents, and $\mathcal{E}_{c}[k]$ is the set of all pairs of agents $(a,b) \in \mathcal{N} \times \mathcal{N}$ that can communicate with each other at time $k$.
Let $\mathbf{M}[k] \in \mathbb{R}^{\it {N} \times \it {N}}$ be the {\it adjacency matrix} with elements $m_{ab}[k] = 1$ if $(a,b) \in \mathcal{E}_c[k]$  and $m_{ab}[k] = 0$ otherwise.  
We define $\mathbf{L}[k] \in \mathbb{R}^{N \times N}$ as the graph Laplacian,  whose elements are $l_{ab}[k] = \sum_{b=1}^{N} m_{ab}[k] = deg(v_a)$ if $a = b$ and $l_{ab}[k] = -m_{ab}[k]$ if $a \neq b$.
Given the agent dynamics \eqref{eqn:DiscreteStateTransitionDynamics} on the spatial grid, each agent $a$ updates its information state at each time $k$ according to a consensus protocol similar to one developed in \cite{ren2008consensus}. This update is based on the agent's current information; the information from all its neighboring agents, of which there are at most $d_{max} = N-1$;
and the reference information state:  
\begin{equation}
\begin{split}
    \xi_{a} [k+1] = \xi_{a}[k]  -
    \alpha \sum_{b \in \mathcal{N}^a_k}  
     l_{ab}[k](\xi_{a}[k] - \xi_{b}[k]) \\ 
      - ~ g_{a} (\xi_{a}[k] - \xi^{r}) , 
      \end{split}
    \label{eqn:ReferenceStateInforamtionDynamicsAgentswise}
\end{equation}
where $a,b \in \mathcal{N}$; $\alpha$ is a constant, chosen such that $\alpha \in (0,\frac{1}{d_{max}})$ \cite{olfati2004consensus}; and $g_{a}$ is defined as:
\begin{equation}
    g_{{a}} = \begin{cases}
    1, &  Z_{k}^{a} \in \mathcal{Z}^{r} \\
    0, & {\rm otherwise}
    \end{cases}
\end{equation}

In the next two sections, we will show that when agents move on the spatial grid according to  \eqref{eqn:DiscreteStateTransitionDynamics} and exchange information with their neighbors according to \eqref{eqn:ReferenceStateInforamtionDynamicsAgentswise}, they achieve {\it average consensus} on their information states, defined as follows: 
\begin{definition}
We say that the vector $\boldsymbol{\xi}[k]$ converges almost surely to average consensus if
\begin{align}
    \boldsymbol{\xi}[k]~ \overset{a.s}{\rightarrow} ~  \xi^{r} \mathbf{1},
\end{align}
\label{def:DirectedGraphAverageConsensus}
where $\mathbf{1} \in \mathbb{R}^{N \times 1}$ is a vector of ones. 
\end{definition}
This implies that the agents' individual information states will eventually converge to a common information state that indicates the presence of the object being searched. We define $\mathcal{T}_{c}$ as the time $k$ at which every agent's information state $\xi_a[k]$ reaches $\xi^{r}$ within a small tolerance $\epsilon$, where $0 \leq \epsilon \ll 1$; i.e., $|\xi_a[\mathcal{T}_{c}] - \xi^r| < \epsilon$ for all agents $a \in \mathcal{N}$. We consider $\mathcal{T}_{c}$ to be the time at which the agents reach consensus.

\begin{algorithm}[t]
\SetAlgoLined
\DontPrintSemicolon
\SetNoFillComment
\textbf{Input:} $\alpha, g_a, \epsilon, \xi^r$; $\xi_a[0] \sim \mathcal{U}(0,1)$; $Z^a_0 \gets i  \in \mathcal{S}$ 
\\
\textbf{Output:} $k, \xi_a[k]$ for which $|\xi_a[k] - \xi^r| \leq \epsilon$ \\
$k \leftarrow 0$ \\

\While{$|\xi_a[k] - \xi^r| > \epsilon$}{
 sum1 $\leftarrow 0$ \\
 sum2 $\leftarrow 0$ \\
\ForAll{$b \in \mathcal{N}^a_k$}
{\tcc{agents $a$, $b$ communicate}
 sum1 $\leftarrow$ sum1  $- ~\alpha l_{ab}[k](\xi_{a}[k] - \xi_{b}[k])$ \\}
\If{$i \in \mathcal{Z}^r$} 
{\tcc{agent $a$ detects feature}
sum2 $\leftarrow - g_a (\xi_{a}[k] - \xi^r)$ \\}
$\xi_{a}[k+1] \leftarrow \xi_{a}[k]$ + sum1 + sum2  \\
$Z_{k+1}^a \leftarrow j$, $(i,j) \in \mathcal{E}_s$, with probability $p_{ij}$ \\
$i \leftarrow j$ \\
$k \leftarrow k+1$ \\
\caption{Control strategy for agent $a \in \mathcal{N}$}
\label{alg:consensus_pseudo_code}
}
\end{algorithm}

The implementation of this probabilistic search strategy on each agent is described in the pseudo code shown in Algorithm \ref{alg:consensus_pseudo_code}. We illustrate the strategy for a scenario with two quadrotors in  \autoref{fig:MotionStrategy_Visual}. The quadrotors start at the spatial grid nodes indexed by $i$ and $j$ and move on the grid according to the DTDS Markov chain dynamics in \eqref{eqn:DiscreteStateTransitionDynamics}. The figure shows sample paths of the quadrotors. The orange quadrotor detects the feature, indicated by a magenta star, when it moves to a node in the set $\mathcal{Z}^r$ (at these nodes, the feature is within the quadrotor's sensing range). 
The quadrotors meet at grid node $m$ after $k=9$ time steps and  exchange information according to \eqref{eqn:ReferenceStateInforamtionDynamicsAgentswise}. They stop the search if their information states are within $\epsilon$ of $\xi^r$; otherwise, they continue to random-walk on the grid.

\section{Analysis of the Markov Chain Model of Agent Mobility}\label{sec:DSMC}
Consider the DTDS  Markov chain that governs the probability mass function of the state $Z_k^{a}$, defined as the location of agent $a$ at time $k$ on the spatial grid that represents the environment.
Then, the time evolution of the agent $a$'s movement in this finite state space can be expressed by using the Markov property as follows:
\begin{equation}
\begin{split}
    Pr(Z_{k+1}^{a} = j ~|~ Z_{k}^{a} = i, Z_{k-1}^{a} = m, \ldots, Z_0^{a} = l ) \\
    = Pr(Z_{k+1}^{a} = j ~|~ Z_k^{a} = i), 
\end{split}
\label{eqn:Markov_property}     
\end{equation}
where the second expression is the probability with which an agent at node $j$ transitions to node $i$ at time $k+1$, and $m,l \in \mathbb{Z}_+$.

\subsection{State Transition Matrix}
The Markov chain \eqref{eqn:DiscreteStateTransitionDynamics} is expressed in terms of the \textit{state transition matrix} $\mathbf{P}$. 
The time invariant matrix $\mathbf{P}$  is defined by the state space of the spatial grid representing the discretized environment. Hence, the Markov chain is $\textit{time-homogeneous}$, which implies that
$Pr(Z^a_{k+1} = j ~|~ Z^a_k = i)$ is the same for all agents at all times $k$.
The entries of $\mathbf{P}$, which are the state transition probabilities, can therefore be defined as
\begin{equation}
    p_{ij}= Pr(Z_{k+1}^{a} = j ~|~ Z_{k}^{a} = i),~ \forall i,j \in \mathcal{S}, ~k \in \mathbb{Z}_+, ~\forall a \in \mathcal{N}.
    \label{eqn:TransitionProbabilityMatHomogenousMC}
\end{equation}

Since each agent chooses its next node from a uniform distribution, these entries can be computed as 
\begin{equation}
         p_{ij} =\begin{cases} 
         \frac{1}{d_{i}+1}, & (i,j) \in \mathcal{E}_{s}, \\
          0, & $otherwise$, 
    \end{cases}
    \label{eqn:TransitionMat_Elements}
\end{equation}
where $d_{i}$ is the degree of the node $i \in \mathcal{S}$, defined as $d_i = 2$ if  $i$ is a corner of the spatial grid, $d_i = 3$ if it is on an edge between two corners, and $d_i = 4$ otherwise. 
Since each entry $p_{ij} \geq 0$, we use the notation $\mathbf{P} \geq 0$. We see that $\mathbf{P}^{m} \geq 0$ for $m \geq 1$. Hence, $\mathbf{P}$ is a \textit{non-negative matrix.} Using Theorem 5 in \cite{grimmett2001probability}, we can conclude that the state transition matrix $\mathbf{P}$ is a stochastic matrix.


\subsection{Stationary Distribution}
A stationary distribution of a Markov chain is defined as follows.
\begin{definition}(Page 227 in \cite{grimmett2001probability})
The vector $\pi \in \mathbb{R}^{S}$ is called a stationary distribution of a  Markov chain if $\pi$ has entries such that:
\begin{enumerate}
    \item $\pi_{j} \geq 0$ $~~\forall j \in \mathcal{S} $
    and $\sum_{j=1}^{S} \pi_{j} =1$ 
    \item $\pi \mathbf{P} = \pi $
\end{enumerate}
\end{definition}

Thus, if $\pi$ is a stationary distribution, we can say that  $\forall k \in \mathbb{Z}_+$, 
\begin{equation}
     \pi \mathbf{P}^{k} = \pi.
    \label{eqn:StationaryDist}
\end{equation}

From the construction of the Markov chain \eqref{eqn:DiscreteStateTransitionDynamics}, each agent has a positive probability of moving from any node $i \in \mathcal{S}$ to any other node $j \in \mathcal{S}$ of the spatial grid in a finite number of time steps.
As a result, the Markov chain $Z_k^{a}$ is an {\it irreducible} Markov chain, and therefore $\mathbf{P}$ is an irreducible matrix. 



From Lemma 8.4.4 (Perron-Frobenius) in \cite{horn1990matrix}, we know that there exists a real unique positive left eigenvector of $\mathbf{P}$. Moreover, since $\mathbf{P}$ is a stochastic matrix, its spectral radius $\rho(\mathbf{P})$ is equal to 1. 
Therefore, we can conclude that this left eigenvector is the stationary distribution of the corresponding DTDS Markov chain. We will next apply the following theorem.

\begin{theorem}
(Theorem 21.12 in \cite{levin2017markov}) An irreducible Markov chain with transition matrix $\mathbf{P}$ is positive recurrent if and only if there exists a probability distribution $\pi$ such that $\pi \mathbf{P} = \pi$.
\label{thm:PositiveRecurecurrenceThm}
\end{theorem}

Since we have shown that the Markov chain  is {\it irreducible} and has a stationary distribution $\pi$, which satisfies $\pi \mathbf{P} = \pi$, we can conclude from Theorem \ref{thm:PositiveRecurecurrenceThm} that the Markov chain is {\it positive recurrent}. 
Thus, all states in the Markov chain are positive recurrent, which implies that each agent will keep visiting every state on the finite spatial grid infinitely often.

\section{Analysis of Consensus on Agents' Information States}\label{sec:ConsensusBasedSearch}
The dynamics of all agents' movements on the spatial grid
can be modeled by a composite Markov chain with states defined as $\mathbf{Z}_k = (Z^1_k, Z^2_k, ..., Z^N_k) \in \mathcal{M}$, where $\mathcal{M} = \mathcal{S}^{\mathcal{N}}$. Note that $S = |\mathcal{S}|$ and $|\mathcal{M}| = S^N$.
We define an undirected graph $\hat{\mathcal{G}} = (\hat{\mathcal{V}},\hat{\mathcal{E}})$ that is associated with the composite Markov chain. 
The vertex set $\hat{\mathcal{V}}$ is the set of all possible realizations $\hat{\imath} \in \mathcal{M}$ of $\mathbf{Z}_k$. The notation $\hat{\imath}(a)$ represents the $a^{th}$ entry of $\hat{\imath}$, which is the spatial node $i \in \mathcal{S}$ occupied by agent $a$.
We define the edge set $\hat{\mathcal{E}}$ of the graph $\hat{\mathcal{G}}$ as follows: $(\hat{\imath},\hat{\jmath}) \in \hat{\mathcal{E}}$ if and only if $(\hat{\imath}(a),\hat{\jmath}(a)) \in \mathcal{E}_s$ for all agents $a \in \mathcal{N}$. Let $\mathbf{Q} \in \mathbb{R}^{ | \mathcal{M}| \times |\mathcal{M}|}$ be the state transition matrix associated with the composite Markov chain.
The elements of $\mathbf{Q}$, denoted by $q_{\hat{\imath} \hat{\jmath}}$, are computed from the transition probabilities defined by Equation \eqref{eqn:TransitionMat_Elements} as follows: 
\begin{equation}
    q_{\hat{\imath} \hat{\jmath}} = \prod_{a=1}^{N} p_{\hat{\imath}(a) \hat{\jmath}(a)}, ~~~~ \forall \hat{\imath}, \hat{\jmath} \in \mathcal{M}. 
    \label{eqn:TransitionRatesProductState}
\end{equation}
In the above expression, $q_{\hat{\imath} \hat{\jmath}}$ is the probability that in the next time step, each agent $a$ will move from spatial node $\hat{\imath}(a)$ to node $\hat{\jmath}(a)$.

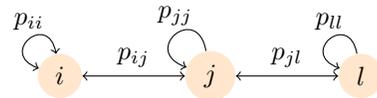
\begin{figure}[h]
    \centering
    \begin{tikzpicture}
        \node[style={circle,fill=orange!20}] (a1) at (0,0) {$i$};
        \node[style={circle,fill=orange!20}] (a2) at (2,0) {$j$};
        \node[style={circle,fill=orange!20}] (a3) at (4,0) {$l$};
        \node[] (a4) at (5,0){};
        \draw[<->,style={draw=black}] (a1) -- node[above]{$p_{i j}$} (a2);
        \draw[<->,style={draw=black}] (a2) -- node[above]{$p_{j l}$} (a3);
        \draw[<->,style={draw=black}] (a1) to [in=160, out=100,looseness=5] node[above] {$p_{i i}$} (a1);
        \draw[->,style={draw=black}] (a2) to [in=160, out=100,looseness=5] node[above] {$p_{j j}$} (a2);
        \draw[->,style={draw=black}] (a3) to [in=160, out=100,looseness=5] node[above] {$p_{l l}$} (a3);
    \end{tikzpicture}
    \caption{A graph $\mathcal{G}_s =  (\mathcal{V}_s,\mathcal{E}_s)$  defined on the set of spatial nodes  $\mathcal{V}_s = \{i,j,l\}$. The arrows signify directed edges between pairs of distinct nodes or self-edges. The edge set of the graph is $\mathcal{E}_s = \{(i,i), (j,j), (l,l), (i,j), (j,l)\}$. }
    \label{fig:MarkovianTransition}
\end{figure} 
\begin{figure}[h]
    \centering
    \begin{tikzpicture}[scale=.5,auto=center]
        \node  [style={circle,fill=green!30!white}] (a1i1a2i1) at (0,0) {$(i,i)$};
        \node (ghatv1) at (0,1.7) {$\hat{i}$};
        \node [style={circle,fill=green!30!white}] (a1i1a2i2) at (5,0) {$(i,j)$};
        \node (ghatv2) at (5,1.7) {$\hat{j}$};
        \node  [style={circle,fill=green!30!white}] (a1i1a2i3) at (10,0) {$(i,l)$};
        \node (ghatv3) at (10,1.7) {$\hat{l}$}; 
        \node (ghatend) at (13,0){}; 
        \draw[<->,style={draw=black}] (a1i1a2i1) -- node[above,midway]{$q_{\hat{i},\hat{j}}$} (a1i1a2i2);
        \draw[<->,style={draw=black}] (a1i1a2i2) -- node[above,midway]{$q_{\hat{j},\hat{l}}$} (a1i1a2i3);
        \draw[style={draw=black, dashed}] (a1i1a2i3) -- (ghatend);
        \draw[<->,style={draw=black}] (a1i1a2i1) to [loop, in=250, out=210,looseness=3] node[below] {$q_{\hat{i},\hat{i}}$} (a1i1a2i1);
        \draw[<->,style={draw=black}] (a1i1a2i2) to [loop, in=250, out=210,looseness=3]
        node[below] {$q_{\hat{j},\hat{j}}$} (a1i1a2i2);
        \draw[->,style={draw=black}] (a1i1a2i3) to [loop, in=250, out=210,looseness=3] 
        node[below] {$q_{\hat{l},\hat{l}}$} (a1i1a2i3);
    \end{tikzpicture}
    \caption{A subset of the composite graph $\mathcal{\hat{G}}=(\mathcal{\hat{V}},\mathcal{\hat{E}})$ for 2 agents that 
    move on the graph $\mathcal{G}_s$ shown in \autoref{fig:MarkovianTransition}.}
    \label{fig:Q_Mat_elem_Vis}
\end{figure}
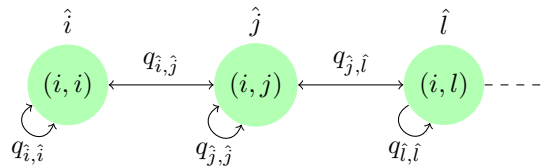

For example, consider a set of two agents, $\mathcal{N} = \{1,2\}$, that move on the graph $\mathcal{G}_s$ as shown in \autoref{fig:MarkovianTransition}.
The agents can stay at their current node in the next time step or travel between nodes $i$ and $j$ and between nodes $j$ and $l$, but they cannot travel between nodes $i$ and $l$.  
\autoref{fig:Q_Mat_elem_Vis} shows a subset of the resulting composite graph $\mathcal{\hat{G}}$.
The set of nodes in the graph $\hat{\mathcal{G}}$ is $\hat{\mathcal{V}} = \{(i,i), (i,j), (i,l), (j,i), (j,j), (j,l), (l,i), (l,j), (l,l)\}$. Each node in $\hat{\mathcal{V}}$ is labeled by a single index $\hat{\imath}$, e.g., $\hat{\imath} = (i,j)$, with $\hat{\imath}(1) = i$ and $\hat{\imath}(2) = j$. Due to the connectivity of the spatial grid defined by $\mathcal{E}_s$, we can for example identify $((i,j),(i,l))$ as an edge in $\hat{\mathcal{E}}$, but not $((i,j),(l,l))$.
Since $N = 2$ and $S = 3$, we have that $|\mathcal{M}| = 3^2 = 9$. For each $\hat{\imath},\hat{\jmath} \in \hat{\mathcal{V}}$,
we can compute the transition probabilities in $\mathbf{Q} \in \mathbb{R}^{9 \times 9}$ from Equation \eqref{eqn:TransitionRatesProductState}as follows:  
\begin{eqnarray}
    q_{\hat{\imath} \hat{\jmath}} & = Pr\left(\mathbf{Z}_{k+1} = \hat{\jmath} ~|~ \mathbf{Z}_{k} = \hat{\imath} \right) = p_{\hat{\imath}(1)\hat{\jmath}(1)}p_{\hat{\imath}(2)\hat{\jmath}(2)}, \nonumber \\
    & \hspace{3cm} ~k \in \mathbb{Z}_+.
     \label{eqn:ProductSpaceTransitionProb}
\end{eqnarray}

We now define $\boldsymbol{\hat{\xi}}[k] = [\xi_1[k] ~~\xi_2[k] ~~\ldots~~ \xi_N[k] ~~\xi^r]^T \in \mathbb{R}^{N+1}$ as an augmented information state vector. The dynamics of information exchange among the agents modeled by
Equation \eqref{eqn:ReferenceStateInforamtionDynamicsAgentswise} can then be represented in matrix form as follows:  
\begin{equation}
    \boldsymbol{\hat{\xi}}[k+1] = \mathbf{H}[k] \boldsymbol{\hat{\xi}}[k], 
    \label{eqn:RefUpdateMatrixNotation}
\end{equation}
 where $\mathbf{H}[k] \in \mathbb{R}^{(N+1) \times (N+1)}$
 is defined as  
 \begin{equation}
 \mathbf{H}[k]=
    \begin{bmatrix}
    ~\mathbf{I} - \alpha \mathbf{L}[k] + \hspace{0.5mm} diag(\mathbf{d}) & ~~ -\mathbf{d} ~ \\
    ~\mathbf{0} & ~~ 1 ~
    \end{bmatrix}
    \label{eqn:RefTrackingUpdateMtrix}
\end{equation}
in which $\mathbf{d} = [g_{1} ~ g_{2} ~ \ldots ~ g_{N}]^T$, $\mathbf{0} \in \mathbb{R}^{1 \times N}$ is a vector of zeros, and $\mathbf{I} \in \mathbb{R}^{N \times N}$ is the identity matrix.   

We associate Equation \eqref{eqn:RefUpdateMatrixNotation} with a graph $\mathcal{G}_{r}[k]$, an expansion of the graph $\mathcal{G}_c[k]$ that includes information flow from the feature nodes $\mathcal{Z}^r$ to agents that occupy these nodes. Here we consider the feature as an additional agent $a_f = N+1$, which remains fixed. 
Let  $\mathcal{G}_{r}[k] = (\mathcal{V}_{r},\mathcal{E}_{r}[k])$ be a directed graph in which  $\mathcal{V}_{r} = \mathcal{N} \cup a_f$, the set of agents and the feature, and $\mathcal{E}_{r}[k] = \mathcal{E}_c[k] ~\cup~ \mathcal{E}_f[k],$ where $\mathcal{E}_f[k]$ is the set of agent-feature pairs $(a,a_f)$ for which $Z_{k}^{a} \in \mathcal{Z}^{r}$ at time $k$.
In this graph, information flows in one direction from the feature nodes to all agents that occupy a feature node on the finite spatial grid at time $k$. 
In addition, information flows bidirectionally between agents that are neighbors at time $k$. 
We now prove the main result of this paper in the following theorem, which shows that all agents will track the reference feature in the environment almost surely and in a distributed fashion.

\begin{theorem}
Consider a group of $N$ agents whose information states evolve according to Equation \eqref{eqn:RefUpdateMatrixNotation}. The information states of all agents will converge to the reference information state $\xi^r$ almost surely.
\label{thm:DirectedMarkovianConsensus}
\end{theorem}

\begin{proof}
Suppose that at an initial time $k_0$, the locations of the $N$ agents on the spatial grid are represented by the node $\hat{\imath} \in \hat{\mathcal{V}}$. Consider another set of agent locations at a future time $k_0 + k$, represented by the node $\hat{\jmath} \in \hat{\mathcal{V}}$. The transition of the agents from configuration $\hat{\imath}$ to configuration $\hat{\jmath}$ in $k$ time steps corresponds to a random walk of length $k$ on the composite Markov chain $\mathbf{Z}_k$ from node $\hat{\imath}$ to node $\hat{\jmath}$. It also corresponds to a random walk by each agent $a$ on the spatial grid from node $\hat{\imath}(a)$ to node $\hat{\jmath}(a)$ in $k$ time steps.
By construction, the graph $\mathcal{G}_s$ is strongly connected and each of its nodes has a self-edge.
Thus, there exists a discrete time $n>0$ such that, for each agent $a$, there exists a random walk on the spatial grid from node $\hat{\imath}(a)$ to node $\hat{\jmath}(a)$ in $n$ time steps. Consequently, there always exists a random walk of length $n$ on the composite Markov chain $\mathbf{Z}_k$ from node $\hat{\imath}$ to node $\hat{\jmath}$.
Therefore, $\mathbf{Z}_k$ is an irreducible Markov chain. All states of an irreducible Markov chain belong to a single communication class. In this case, all states are \textit{positive recurrent}.
As a result, each state of $\mathbf{Z}_k$ is visited infinitely often by the group of agents. Moreover, because the composite Markov chain is irreducible, we can conclude that $\cup_{k \in \mathbb{Z}_+} \mathcal{G}_c[k] = \mathcal{G}_0$, where $\mathcal{G}_0$ is the complete graph on the set of agents 
$\mathcal{N}$, and therefore that $\cup_{k \in \mathbb{Z}_+} \mathcal{G}_r[k]$ contains a directed spanning tree with 
$\xi^{r}$ as the fixed root. Since this union of graphs has a spanning tree, we can apply Theorem 3.1 in \cite{matei2009consensus} to conclude that the information state of each agent will converge to $\xi^{r}$ almost surely. The notation $\theta(k)$ and $F_{\theta(k)}$ in \cite{matei2009consensus} corresponds to our definitions of $\mathbf{Z}_k$ and $\mathbf{H}[k]$, respectively. \qed 
\end{proof}

\section{Simulation Results}\label{sec:Sim}
We validate the result on average information consensus in Theorem \ref{thm:DirectedMarkovianConsensus} with numerical simulations in MATLAB and 3D physics-based
software-in-the-loop (SITL) simulations developed in ROS-Melodic and Gazebo 9 \cite{koenig2004design}. In the simulations, multiple agents perform random walks  on a finite spatial grid according to the dynamics in Equation \eqref{eqn:DiscreteStateTransitionDynamics}. Each grid is defined as a square lattice with $c = \sqrt{S}$
nodes on each side, where the distance between neighboring nodes is $d = 1$ m.
The state transition probabilities $p_{ij}$ of the corresponding graph $\mathcal{G}_s$ are defined according to Equation \eqref{eqn:TransitionMat_Elements}. Since our largest simulated agent population is $N=14$, and the parameter $\alpha$ must be less than $\frac{1}{d_{max}} = \frac{1}{N-1}$ \cite{olfati2004consensus}, we set $\alpha = \frac{1}{14-1} \approx 0.08$. The tolerance $\epsilon$ defining the time until consensus was set to 0.01.
All simulations were run on a  desktop computer with 16 GB of RAM and an Intel Xeon 3.0 GHz 16 core processor with an NVIDIA Quadro M4000 graphics processor.  

\subsection{Numerical Simulations}
We performed large ensembles of Monte Carlo simulations to investigate the effect of the number of agents $N$, the spatial grid dimension $c$, and the resulting agent density $N/c^2$ on the expected time until the agents reach consensus, i.e., agree that the feature of interest is present. Quantifying the effect of these factors  is necessary in order to determine the number of agents that should search a given area. This would help first responders to optimally distribute resources for searching a disaster-affected environment.

Each agent is modeled as a point mass that can move between adjacent nodes on the graph $\mathcal{G}_s$, as illustrated in \autoref{fig:MotionStrategy_Visual}. We assume that the agents can localize on $\mathcal{G}_s$. The set of neighbors $\mathcal{N}^a_k$ of an agent $a$ at time $k$ consists of all agents that occupy the same spatial node as agent $a$ at that time. The 
feature can by detected by an agent located at nodes $\mathcal{Z}^r = \{ 4, 5, 6\}$ of the spatial grid, and the reference information state of the feature is defined as $\xi^r = 1$. 

\begin{figure}[t]
    \centering
    \includegraphics[scale=0.55]{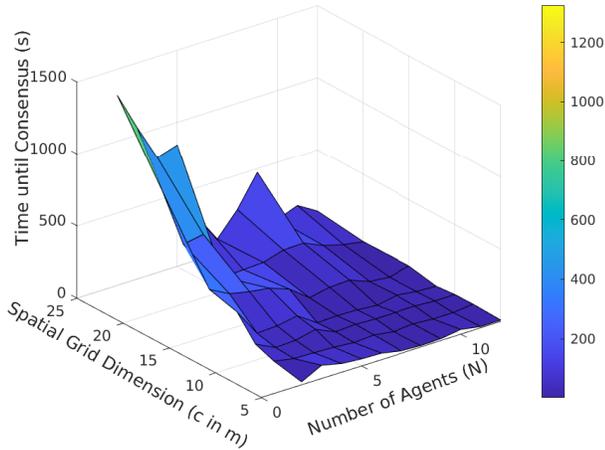}
    \caption{Mean time (s) until consensus is reached, $\mu$, versus number of agents $N$ and spatial grid dimension $c$. Each value of $\mu$ is averaged over 1000 Monte Carlo simulations of scenarios with the corresponding values of $N$ and $c$.}
    \label{fig:SurfacePlotConsensus}
\end{figure}

\begin{figure}[t]
    \centering
    \includegraphics[scale=0.55]{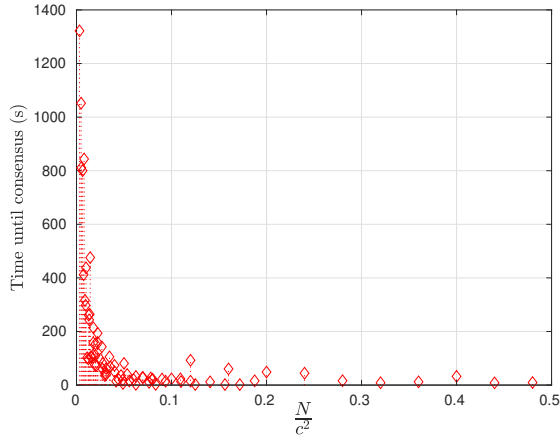}
    \caption{Mean time (s) until consensus is reached, $\mu$, versus agent density $N/c^2$ for the simulation data plotted in \autoref{fig:SurfacePlotConsensus}.}
    \label{fig:mu_vs_agent_density}
\end{figure}

To investigate the dependence of the expected time to reach consensus, $\mathbb{E}[\mathcal{T}_c]$, on the number of agents $N$ and the spatial grid dimension $c$, we simulated scenarios with different combinations of $N \in \{2, 3, \ldots, 14\}$ and $c \in \{5, 8, 10, 12, 15, 20\}$ meters. For each scenario, we ran 1000 simulations with random initial agent positions and computed the mean time $\mu$ at which the agents reached consensus. 
\autoref{fig:SurfacePlotConsensus} plots the  values of $\mu$ versus $N$ and $c$ for each simulated scenario, and \autoref{fig:mu_vs_agent_density} plots $\mu$ versus the corresponding agent density, $N/c^2$. We observe from these figures 
that a decrease in the agent density results in an increase in $\mu$. This can be attributed to low agent encounter rates with other agents and with feature nodes at low agent densities. Using the curve fitting toolbox in MATLAB and data from \autoref{fig:mu_vs_agent_density} we see that there is an exponential relation between $\mathbb{E}[\mathcal{T}_c]$ and $N/c^2$ given by $\mathbb{E}[\mathcal{T}_c] =ae^{-b\frac{N}{c^2}}$ with $a=0.008,b=-15.84$. \autoref{fig:mu_vs_agent_density} shows that the expected time until consensus does not decrease appreciably for agent densities above approximately $N/c^2 = 0.05$. Thus, for a given grid size $c^2$, it may not be necessary to deploy more than about $\lceil 0.05c^2 \rceil$ agents ($0.05c^2$ rounded up to the next integer) to search the area.

\begin{figure}[h!]
    \begin{subfigure}{0.5\textwidth}
        \centering
        \includegraphics[trim = 0.2cm 0 0 0, width=\textwidth,height=0.6\textwidth]{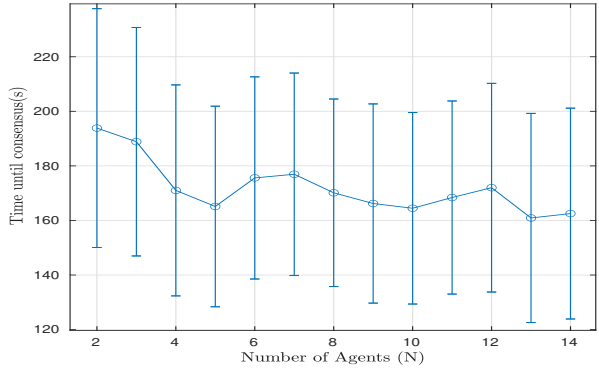}
        \caption{}
        \label{fig:Temporal_mean_with_agent_size}
    \end{subfigure}
    \begin{subfigure}{0.5\textwidth}
        \centering
        \includegraphics[trim = 0.2cm 0 0 0, width=\textwidth,height=0.6\textwidth]{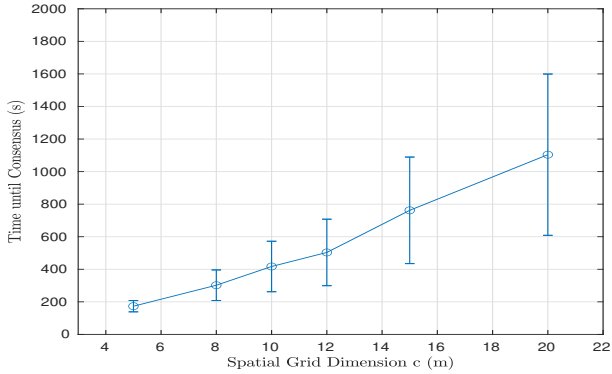}
        \caption{}  
        \label{fig:Temporal_mean_with_grid_size}
    \end{subfigure}
    \caption{Time until consensus is reached, averaged over 1000 Monte Carlo simulations of scenarios with
    (a) varying numbers of agents $N$ and grid dimension $c = 5$; (b) varying $c$ and $N = 5$. The circles mark mean times $\mu$, and the error bars show standard deviations $\sigma$.}
    \label{fig:consensusTimeVariation}
\end{figure}

For selected combinations of $N$ and $c$, we also computed the standard deviation $\sigma$ of the time to reach consensus over the corresponding 1000 simulations.
\autoref{fig:Temporal_mean_with_agent_size} plots $\mu \pm \sigma$ versus $N$ for a fixed grid dimension $c = 5$, and \autoref{fig:Temporal_mean_with_grid_size} plots $\mu \pm \sigma$ versus $c$ for a fixed number of agents $N=5$. \autoref{fig:Temporal_mean_with_agent_size} shows that for a relatively small grid size ($c=5$), both $\mu$ and $\sigma$ do not vary substantially with $N$. Thus, a small number of agents would be sufficient to search such an environment, since increasing the agent density would not significantly speed up the search or reduce the variability in time until consensus.
\autoref{fig:Temporal_mean_with_grid_size} indicates that for a fixed group size of $N=5$ agents, both $\mu$ and $\sigma$ increase monotonically with the size of the grid. This trend suggests that more agents should be deployed if the predicted time until consensus and/or the variability in this time is too high for a given environment.

\begin{figure}[t]
    \begin{subfigure}{0.5\textwidth}
        \centering
        \includegraphics[width=\textwidth,height=0.5\textwidth]{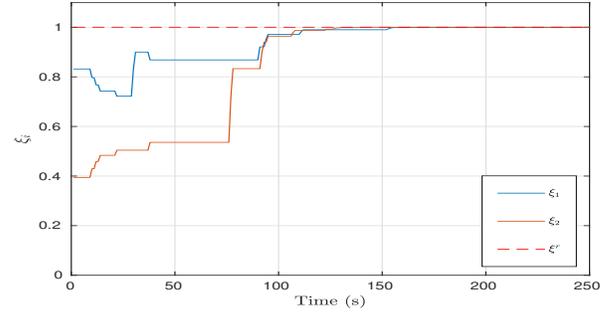} 
        \caption{}
        \label{fig:StateTrackingConsensus2N_3X3}
    \end{subfigure}
    \begin{subfigure}{0.5\textwidth}
        \centering
        \includegraphics[width=\textwidth,height=0.5\textwidth]{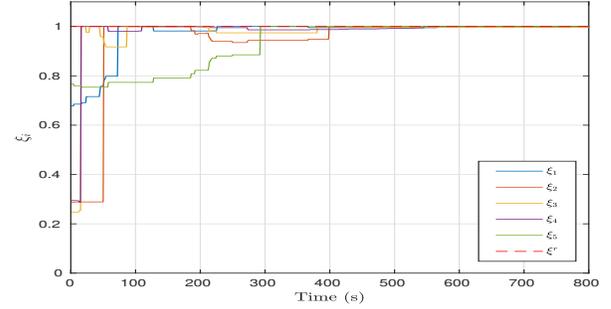}
        \caption{}
        \label{fig:StateTrackingConsensus5N_10X10}
    \end{subfigure}
    \caption{Time evolution of the agent information states $\xi_a[k]$ in simulations of (a) $N=2$ agents moving on a 3$\times$3 grid; (b) $N=5$ agents moving on a 10$\times$10 grid.}
    \label{fig:consenus_plots_matlab}
\end{figure}

We illustrate the agents' consensus dynamics with two cases of the simulation runs. Figure \ref{fig:consenus_plots_matlab}  plots the time evolution of the agent information states for each case. In the first case, $N =2$ agents traverse a spatial grid with dimension $c=3$. From \autoref{fig:StateTrackingConsensus2N_3X3}, we see that the time until consensus, i.e. the time at which both agents' information states converge within $\epsilon$ of the reference state $\xi^r = 1$, is approximately 160 s. We also simulate $N = 5$ agents that traverse a  spatial grid with dimension $c=10$. \autoref{fig:StateTrackingConsensus5N_10X10} shows that the time until consensus has increased to about 570 s in this case, which is  within one standard deviation $\sigma$ of the mean consensus time $\mu$ computed from our Monte Carlo analysis, as shown in \autoref{fig:Temporal_mean_with_grid_size} for $c=10$.

We also studied  the effect on $\mathbb{E}[\mathcal{T}_c]$ of uncertainty in the agents' identification of the feature nodes (i.e., $\xi^r$ is a random variable), which may arise in practice due to factors such as sensor noise, occlusion of features, and inter-agent communication failures. We ran 1000 Monte Carlo simulation runs, for each of two scenarios, all with $N=5$ agents moving on a spatial grid with dimension $c = 5$ m. For each scenario, \autoref{tab:Consensus_MC_analysis2} shows the mean $\mu$ and standard deviation $\sigma$ of the time until the agents reach consensus.
To investigate the effect of uncertainty in feature identification, we specified that agents either perfectly identify the feature, in which case $\xi^r = 1$, or obtain noisy measurements of the feature, for which  $\xi^{r} \sim \mathrm{N}(1,0.02)$.
From \autoref{tab:Consensus_MC_analysis2}, we observe that the addition of noise to the agents' measurements of the feature results in an increase in both $\mu$ and $\sigma$. However, despite information uncertainty, the agents successfully achieve consensus. 

\begin{table}[t]
\centering
\begin{tabular}{| c | c | c |}
    \hline
      Reference information & Time until consensus is reached (s) \\
    state &  $\mu \pm \sigma$  \\
    \hline
    $\xi^{r} =1$ & 140 $\pm$ 35 \\
    \hline
    $\xi^{r} \sim \mathrm{N}(1,0.02)$ & 175 $\pm$ 68 \\
    \hline
\end{tabular}
\caption{Time until consensus is reached ($\mu \pm \sigma$), computed from 1000 Monte Carlo simulations of scenarios with $N=5$, $c=5$ and different values of $\xi^r$}
\label{tab:Consensus_MC_analysis2}
\end{table}


\begin{figure}[t!]
    \centering
    \begin{subfigure}{0.5\textwidth}
    \includegraphics[height=0.5\textwidth, width=\textwidth]{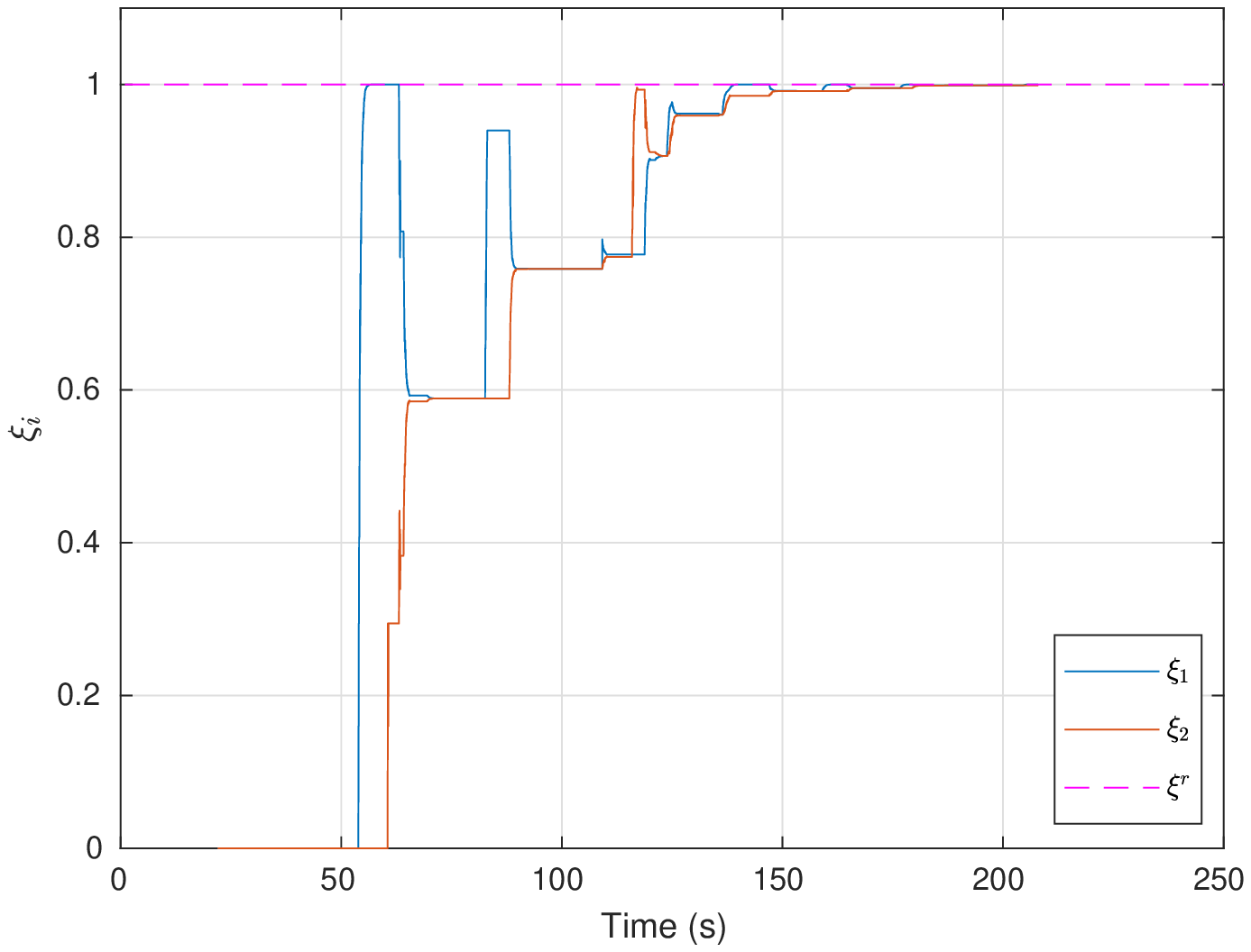}
    \caption{}
    \label{fig:Gazebo_sim_2_AG_3x3}
    \end{subfigure}
    \begin{subfigure}{0.5\textwidth}
    \includegraphics[height=0.5\textwidth, width=\textwidth]{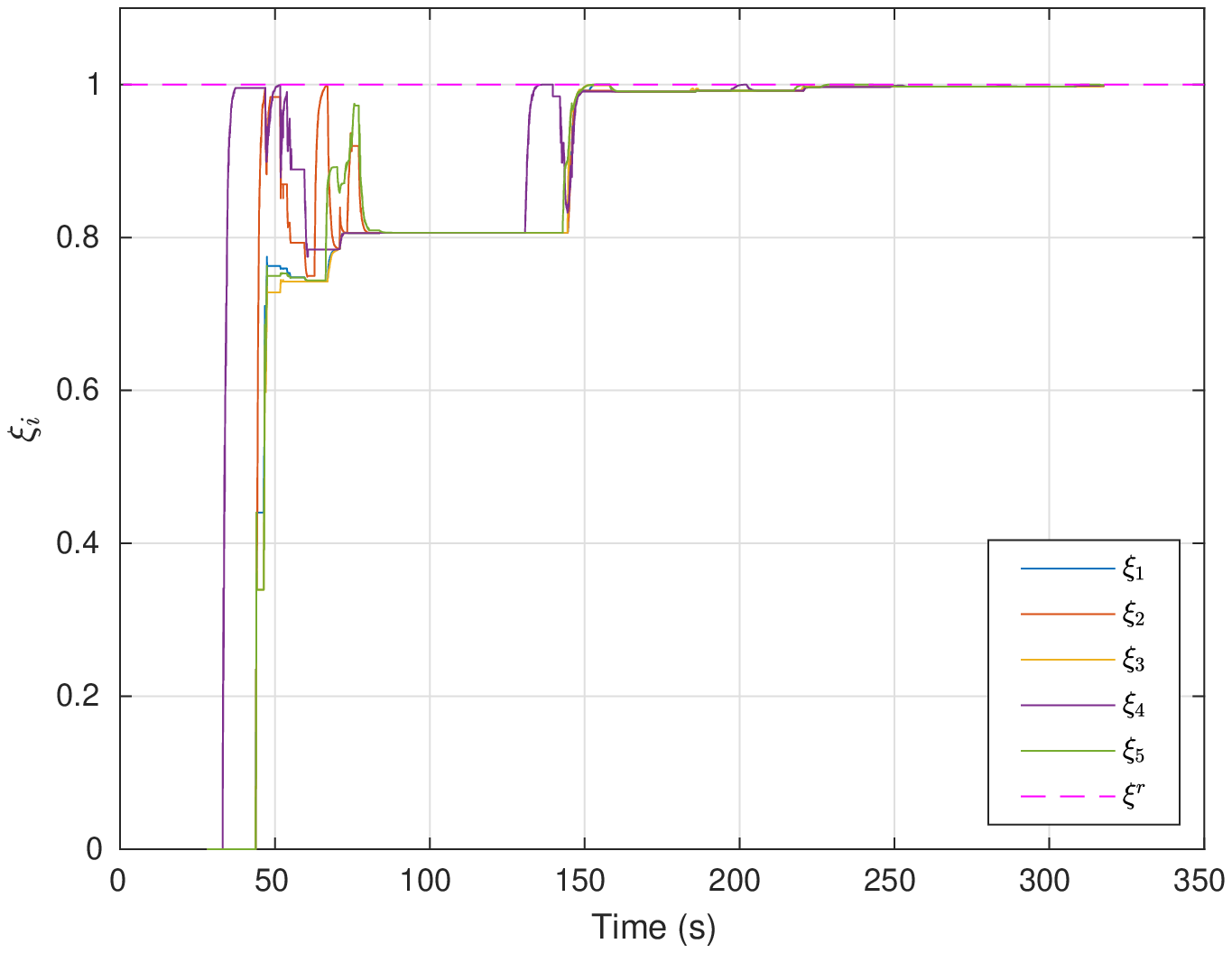}
    \caption{}
    \label{fig:Gazebo_sim_5_AG_5x5}
    \end{subfigure}
    \caption{Time evolution of the robot information states $\xi_a[k]$ in  Gazebo simulation runs of (a) $N=2$ robots moving on a 3$\times$3 grid; (b) $N=5$ robots moving on a 5$\times$5 grid.}
    \label{fig:Gazebosims}
\end{figure}

\subsection{3D Physics Simulations}
We also tested our search strategy in physics-based simulations. A snapshot of the Gazebo simulation environment is shown in \autoref{fig:gazebo_environment}. The agents are modeled as quadrotors with a plus frame configuration. We assume that the agents can accurately localize in the environment using onboard inertial and GPS sensors. The analysis of our probabilistic consensus strategy under localization uncertainty is beyond the scope of this paper. We also assume that the feature of interest is known to be present in the environment, but its location is unknown. 
 
Each quadrotor is equipped with a downward-facing RGB camera with a resolution of $1080 \times 720$. The feature of interest is modeled as a magenta box, which the agents detect from their camera images using a color-based classifier. We added zero-mean Gaussian noise with standard deviation $0.07$ to the photometric intensity in the camera sensor model. We also used a standard plumb bob distortion model to account for camera lens distortion.  The quadrotors are spaced 0.5 m apart in altitude in order to prevent collisions. The altitude difference causes slight disparities in the quadrotors' field-of-view (FOV), but this does not significantly affect the performance of the search strategy.

We simulated two scenarios:  $N=2$ robots at altitudes $0.5$ m and $1$ m traversing a $3 \times 3$ grid, and $N=5$ robots at altitudes between $1$ m and $3$ m traversing a $5 \times 5$ grid. The video attachment (also online at \url{https://youtu.be/j74jeWQ0HM0}) shows a simulation run of the second scenario. \autoref{fig:Gazebo_sim_2_AG_3x3} and \autoref{fig:Gazebo_sim_5_AG_5x5} plot the time evolution of the agent information states over a single simulation run of each scenario. The information states sometimes display steep drops in value, as in the plots of $\xi_2$ and $\xi_4$ in \autoref{fig:Gazebo_sim_5_AG_5x5} from 50 s to 70 s. These drops can be attributed to the following factors: (1) an agent updates its information state with states communicated by its neighbors, according to the consensus protocol; (2) an agent that is at the feature node stops detecting the feature below when another agent at a lower altitude enters its field of view, occluding the feature; (3) spurious measurements like false positives may have been introduced by an agent's sensors. Despite the unmodeled effects of the second and third factors on the information states, the agents still successfully reach consensus during the Gazebo simulations.  We see that the time until consensus is reached in \autoref{fig:Gazebo_sim_2_AG_3x3} and \autoref{fig:Gazebo_sim_5_AG_5x5}
is about 210 s and 250 s, respectively. The delays in these times compared to the times in the Monte Carlo simulations in \autoref{fig:consensusTimeVariation} can be attributed to the second and third factors described above and to the inertia of the quadrotor, which affect the Gazebo simulations but not the Monte Carlo simulations.
\section{Conclusion and Future Work}\label{sec:ConcFutre}
In this paper, we have presented a probabilistic search strategy for multiple agents with local sensing and communication capabilities. The agents explore a bounded environment according to a DTDS Markov motion model and share information with  neighboring agents. We proved that the agents achieve consensus almost surely on the presence of a static feature of interest in the environment. Thus, agents that do not detect the feature through direct measurement will eventually recognize its presence through information exchange with other agents. Importantly, this result does not require any assumptions on the connectivity of the agents' communication network. Thus, the search strategy is suitable for applications in which network connectivity is difficult to maintain, such as disaster scenarios.
We investigated the performance of our strategy in both numerical and physics-based simulations.  

In future work, we will extend this probabilistic search strategy to enable the agents to localize the target(s) in the environment using distributed consensus methods. We also plan to experimentally implement our strategy on 
aerial robots equipped with RGB-D cameras and utilize
more robust feature classifiers, such as  SURF \cite{bay2008speeded}, for identification of features. We propose to use  
5G WiFi modules for inter-robot communication in order to facilitate high-bandwidth data exchange.

\bibliography{main}

\begin{thebibliography}{10}

\bibitem{michael2014collaborative}
Nathan Michael, Shaojie Shen, Kartik Mohta, Vijay Kumar, Keiji Nagatani,
  Yoshito Okada, Seiga Kiribayashi, Kazuki Otake, Kazuya Yoshida, Kazunori
  Ohno, et~al.
\newblock Collaborative mapping of an earthquake damaged building via ground
  and aerial robots.
\newblock In {\em Field and Service Robotics}, pages 33--47. Springer, 2014.

\bibitem{burgard2005coordinated}
Wolfram Burgard, Mark Moors, Cyrill Stachniss, and Frank~E Schneider.
\newblock Coordinated multi-robot exploration.
\newblock {\em IEEE Transactions on Robotics}, 21(3):376--386, 2005.

\bibitem{nagatani2013emergency}
Keiji Nagatani, Seiga Kiribayashi, Yoshito Okada, Kazuki Otake, Kazuya Yoshida,
  Satoshi Tadokoro, Takeshi Nishimura, Tomoaki Yoshida, Eiji Koyanagi, Mineo
  Fukushima, et~al.
\newblock Emergency response to the nuclear accident at the {Fukushima Daiichi
  Nuclear Power Plants} using mobile rescue robots.
\newblock {\em Journal of Field Robotics}, 30(1):44--63, 2013.

\bibitem{simmons2000coordination}
Reid Simmons, David Apfelbaum, Wolfram Burgard, Dieter Fox, Mark Moors,
  Sebastian Thrun, and H{\aa}kan Younes.
\newblock Coordination for multi-robot exploration and mapping.
\newblock In {\em AAAI/IAAI}, pages 852--858, 2000.

\bibitem{koenig2004design}
Nathan Koenig and Andrew Howard.
\newblock Design and use paradigms for gazebo, an open-source multi-robot
  simulator.
\newblock In {\em 2004 IEEE/RSJ International Conference on Intelligent Robots
  and Systems (IROS)}, volume~3, pages 2149--2154. IEEE, 2004.

\bibitem{howard2006experiments}
Andrew Howard, Lynne~E Parker, and Gaurav~S Sukhatme.
\newblock Experiments with a large heterogeneous mobile robot team:
  Exploration, mapping, deployment and detection.
\newblock {\em The International Journal of Robotics Research},
  25(5-6):431--447, 2006.

\bibitem{husain2013mapping}
Ammar Husain, Heather Jones, Balajee Kannan, Uland Wong, Tiago Pimentel, Sarah
  Tang, Shreyansh Daftry, Steven Huber, and William~L Whittaker.
\newblock Mapping planetary caves with an autonomous, heterogeneous robot team.
\newblock In {\em 2013 IEEE Aerospace Conference}, pages 1--13. IEEE, 2013.

\bibitem{spanos2005dynamic}
Demetri~P Spanos, Reza Olfati-Saber, and Richard~M Murray.
\newblock Dynamic consensus on mobile networks.
\newblock In {\em IFAC World Congress}, pages 1--6. Citeseer, 2005.

\bibitem{ren2007information}
Wei Ren, Randal~W Beard, and Ella~M Atkins.
\newblock Information consensus in multivehicle cooperative control.
\newblock {\em IEEE Control Systems}, 27(2):71--82, 2007.

\bibitem{ren2004consensus}
Wei Ren and Randal~W Beard.
\newblock Consensus of information under dynamically changing interaction
  topologies.
\newblock In {\em 2004 American Control Conference}, volume~6, pages
  4939--4944. IEEE, 2004.

\bibitem{mesbahi2010graph}
M~Mesbahi and M~Egerstedt.
\newblock {\em Graph theoretic methods in multiagent systems}.
\newblock Princeton University, Princeton, NJ, 2010.

\bibitem{olfati2004consensus}
Reza Olfati-Saber and Richard~M Murray.
\newblock Consensus problems in networks of agents with switching topology and
  time-delays.
\newblock {\em IEEE Transactions on Automatic Control}, 49(9):1520--1533, 2004.

\bibitem{parasuraman2018multipoint}
Ramviyas Parasuraman, Jonghoek Kim, Shaocheng Luo, and Byung-Cheol Min.
\newblock Multipoint rendezvous in multirobot systems.
\newblock {\em IEEE Transactions on Cybernetics}, 50(1):310--323, 2018.

\bibitem{yu2020synthesis}
Xi~Yu and M~Ani Hsieh.
\newblock Synthesis of a time-varying communication network by robot teams with
  information propagation guarantees.
\newblock {\em IEEE Robotics and Automation Letters}, 5(2):1413--1420, 2020.

\bibitem{fox2006distributed}
Dieter Fox, Jonathan Ko, Kurt Konolige, Benson Limketkai, Dirk Schulz, and
  Benjamin Stewart.
\newblock Distributed multirobot exploration and mapping.
\newblock {\em Proceedings of the IEEE}, 94(7):1325--1339, 2006.

\bibitem{vincent2008distributed}
Regis Vincent, Dieter Fox, Jonathan Ko, Kurt Konolige, Benson Limketkai, Benoit
  Morisset, Charles Ortiz, Dirk Schulz, and Benjamin Stewart.
\newblock Distributed multirobot exploration, mapping, and task allocation.
\newblock {\em Annals of Mathematics and Artificial Intelligence},
  52(2-4):229--255, 2008.

\bibitem{hu2012multiagent}
Jinwen Hu, Lihua Xie, Kai-Yew Lum, and Jun Xu.
\newblock Multiagent information fusion and cooperative control in target
  search.
\newblock {\em IEEE Transactions on Control Systems Technology},
  21(4):1223--1235, 2012.

\bibitem{martinez2012brownian}
Fredy Martinez, Edwar Jacinto, and Diego Acero.
\newblock Brownian motion as exploration strategy for autonomous swarm robots.
\newblock In {\em 2012 IEEE International Conference on Robotics and
  Biomimetics (ROBIO)}, pages 2375--2380. IEEE, 2012.

\bibitem{wagner1998robotic}
Israel~A Wagner, Michael Lindenbaum, and Alfred~M Bruckstein.
\newblock Robotic exploration, {B}rownian motion and electrical resistance.
\newblock In {\em International Workshop on Randomization and Approximation
  Techniques in Computer Science}, pages 116--130. Springer, 1998.

\bibitem{winfield2000distributed}
Alan~FT Winfield.
\newblock Distributed sensing and data collection via broken ad hoc wireless
  connected networks of mobile robots.
\newblock In {\em Distributed Autonomous Robotic Systems 4}, pages 273--282.
  Springer, 2000.

\bibitem{RageshTRO2020}
Ragesh~K. {Ramachandran}, Zahi {Kakish}, and Spring {Berman}.
\newblock Information correlated {L\'e}vy walk exploration and distributed
  mapping using a swarm of robots.
\newblock {\em IEEE Transactions on Robotics}, 2020.

\bibitem{ren2008consensus}
Wei Ren and Randal~W Beard.
\newblock Consensus tracking with a reference state.
\newblock {\em Distributed Consensus in Multi-vehicle Cooperative Control:
  Theory and Applications}, pages 55--73, 2008.

\bibitem{grimmett2001probability}
Geoffrey Grimmett and David Stirzaker.
\newblock {\em Probability and random processes}.
\newblock Oxford University Press, 2001.

\bibitem{horn1990matrix}
Roger~A Horn and Charles~R Johnson.
\newblock {\em Matrix analysis}.
\newblock Cambridge University Press, 1990.

\bibitem{levin2017markov}
David~A Levin and Yuval Peres.
\newblock {\em Markov chains and mixing times}, volume 107.
\newblock American Mathematical Society, 2017.

\bibitem{matei2009consensus}
Ion Matei, Nuno~C Martins, and John~S Baras.
\newblock Consensus problems with directed {Markovian} communication patterns.
\newblock In {\em 2009 American Control Conference}, pages 1298--1303. IEEE,
  2009.

\bibitem{bay2008speeded}
Herbert Bay, Andreas Ess, Tinne Tuytelaars, and Luc Van~Gool.
\newblock Speeded-up robust features {(SURF)}.
\newblock {\em Computer Vision and Image Understanding}, 110(3):346--359, 2008.

\end{thebibliography}
\end{document}